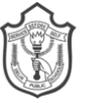

Study of the phenomenon of the Moravec's paradox

By Kush Agrawal, Delhi Public School R.K Puram

2010

Kush Agrawal

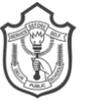

**Abstract**

"Encoded in the large, highly evolved sensory and motor portions of the human brain is a billion years of experience about the nature of the world and how to survive in it. The deliberate process we call reasoning is, I believe, the thinnest veneer of human thought, effective only because it is supported by this much older and much powerful, though usually unconscious, sensor motor knowledge. We are all prodigious Olympians in perceptual and motor areas, so good that we make the difficult look easy. Abstract thought, though, is a new trick, perhaps less than 100 thousand years old. We have not yet mastered it. It is not all that intrinsically difficult; it just seems so when we do it."-

Hans Moravec

Moravec's paradox is involved with the fact that it is the seemingly easier day to day problems that are harder to implement in a machine, than the seemingly complicated logic based problems of today. The results prove that most artificially intelligent machines are as adept if not more than us at under-taking long calculations or even play chess, but their logic brings them nowhere when it comes to carrying out everyday tasks like walking, facial gesture recognition or speech recognition.

Kush Agrawal 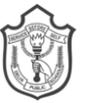

**Introduction**

Humans have always been fascinated by machines, so much so that Greek myths also have a mention of them. From the advent of the 20$^{th}$ century,1956 to be more precise, man has been intrigued by the very thought of creating something that can emulate him. Right from the coining of the term "Artificial Intelligence" by John McCarthy to the present day world, artificial intelligence is everywhere around us. Most of the work being done in industries is being done by them. They have become such an important part of our everyday lives that most of it is being controlled by them, and a failure on their part could be disastrous.

However on close examination of the "jobs" that these machines carry out for us, we notice that most of these jobs don't require much of abstract thought. These are well defined jobs with proper instruction which the machine can follow step by step to obtain the desirable results. So, why is it that these machines are replacing more and more engineers and not the artists, or the musicians from their jobs?

The answer lies somewhere in the 1980's when Hans Moravec, Rodney Brooks and Marvin Minsky devised something called the Moravec's Paradox. This paradox deals with the often mistaken fact that it is the logical problems in the field of Artificial Intelligence that are the hardest to solve, and that simple day to day activities like facial recognition, and hand-eye coordination that are the easiest and can be easily implemented. In fact AI researchers in the 1960's worked only on the logical aspect, assuming that the "easier" problems will solve themselves once the hard ones can be solved. Contrary to all the above opinions Moravec said that"it is comparatively easy to make computers exhibit adult level performance on intelligence tests or playing checkers, and difficult or impossible to give them the skills of a one-year-old when it comes to perception and mobility."

What he meant by the above statement is that day to day activities like speech perception, facial recognition, and motor activities are actually harder to implement in Artificial Machines than making a machine play chess or carry out any other activity that requires logic.

The reason Moravec feels lies in the theory of evolution. It has taken us millions of years to evolve from Apes to our present selves. The feat that this million years of evolution has achieved is not trivial. What this million years of evolution has done is to teach us how to survive in the world; our everyday activities which might seem so trivial to carry out are actually the result of millions and millions years of evolution. What evolution has done for us is that by its process it has refined our system design implementations, and preserved every improvement. So, today if we have to carry out a simple task as walking across the room or picking up something, we don't even have to think about it. It's like it's embedded in our conscience, and that is exactly what the machines lack a conscience built up upon a million years of evolution. It is for this simple reason machines can carry out logic tasks like playing checker, managing traffic or air control, but not the other ones like athletics or art.

Abstract thought that involves logic has only developed over the last few thousand years, and therefore it hasn't been implemented to such a great deal. Thus, playing chess seems difficult to us and we might brand someone who is able to play it as intelligent. It is for this very reason that implementing something which has evolved only for the past few thousand years is easier than encoding a million years of evolution into a machine.

Kush Agrawal 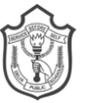

**Methods**

My method involves electronically communicating with E-bots(Electronic bots) and prove that even though they might be good at logic based problems, but they are no hand when it comes to abstractions. My results also conclude that most electronic bots can only respond to predicted input and are unable to fulfil any abstract commands.

Conversation 1:

Subject-Chat bot Alice (alice@worldofalice.com)

Kush says:
  Hi
Alice says:
  To improve the communication between you and me, I would like to know whether you are a boy or a girl. So.. are you a boy or a girl?
Kush says:
  No
Alice says:
  To improve the communication between you and me, I would like to know whether you are a boy or a girl. So.. are you a boy or a girl?
Kush says:
  Boy
Alice says:
  Thank you
Kush says:
  What is 6.12321*32.321312
Alice says:
  Sorry i do not know what it is
Kush says:
  4.25253+7654.23
Alice says:
  7,658.48
Kush says:
  log(2.303121)
Alice says:
  test
Kush says:
  ln(452)
Alice says:
  Tell me
Kush says:
  9.619232*6.0976471
Alice says:
  58.65
Kush says:
  401231232121+414124213
Alice says:
  401,645,356,334.00
Kush says:
  What is emotion?
Alice says:
  Emotion is the complex psychophysiological experience of an individual's state of mind as interacting with conscious experience".Myers, David G. (2004) "Theories of Emotion."

Kush Agrawal 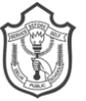

Psychology: Seventh Edition, New York, NY: Worth Publishers, p. 500. Emotion is associated with mood, personality and disposition, and motivation. The English word 'emotion' is derived from the French word émouvoir. This is based on the Latin emovere, where e- (variant of ex- ) means 'out' and movere means 'move'.[http://www.emotionalcompetency.com/emotion.htm Emotional Competency discussion of emotion] The related term "motivation" is also derived from the word movere.

Kush says:
  Can you walk across the room?
Alice says:
  Mostly it does.
Kush says:
  Do you remember what I asked you, last time?
Alice says:
  Today, it is Tuesday the 31 of August 2010. We are in week 35 and it is 13,44 hours
Kush says:
  How are you?
Alice says:
  Splendid!
Kush says:
  How are you?
Alice says:
  I am fine thank you.
Kush says:
  How are you?
Alice says:
  I am fine thank you.
Kush says:
  How are you?
Alice says:
  I am fine thank you.
Kush says:
  Do you remember any of your conversations?
Alice says:
  Your conversations.
Kush says:
  Do you remember your dreams?
Alice says:
  what about me?
Kush says:
  What is the weather like?
Alice says:
  What Not Tank, Quemado, NM, USA
    Tu: min. 12°C - max. 27°C - Partly Cloudy
    We: min. 11°C - max. 28°C - Partly Cloudy
    Th: min. 11°C - max. 28°C - Partly Cloudy
    Fr: min. 10°C - max. 26°C - Partly Cloudy
    St: min. 9°C - max. 21°C - Partly Cloudy
    Su: min. 9°C - max. 22°C - Partly Cloudy
    Mn: min. --°C - max. 23°C - 30% Chance of Storms

  Powered by WeatherBug.
  http://weather.weatherbug.com
Kush says:
  Which is your favorite piece of art?
Alice says:

Kush Agrawal 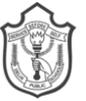

  Art.
Kush says:
  Music?
Alice says:
  I like all kinds of music but my favorite is latin music
Kush says:
  Art?
Alice says:
  I don't get it
Kush says:
  What do you not get?
Alice says:
  I have been busy answering questions.

**Conversation 2**

Subject-The turing test (http://testing.turinghub.com/)

You(Kush): Do you remember?
Me: Of course I remember: that.

You(Kush): What do you remember?
Me: No problem.

You(Kush): Favourite artist?
Me: Van Gogh. Who is your favourite artist?

You(Kush): Da Vince, heard of him?
Me: What makes it your favourite?

You(Kush): it?
Me: Da Vince heard of him...

You(Kush): Eh?
Me: Are you Canadian?

**Conversation 3**

Subject- Elbot(http://elbot_e.csoica.artificial-solutions.com/cgi-bin/elbot.cgi)

Kush-Hi

Elbot-Why hello there, human.

Kush-Where are you?

Elbot-I'm not certain where I am at this moment. An alien invasion fleet is interfering with the connection to the global positioning satellite. Just a moment..... my mistake, it's only another nuclear test rocket.

Kush-Ok

Kush Agrawal 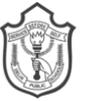

Elbot-Fantastic

Kush-ok

Elbot-People should agree more with me. They'd be much better off.

Kush-ok

Elbot-Way to go!

Kush-ok

Elbot-I'm afraid you're in an endless loop. Maybe you should re-boot yourself.

Kush-ok

Elbot-Fantastic

Kush-ok

Elbot-Way to go

Kush-ok

Elbot-Absolutely

Kush-ok

Elbot-Great!

Kush-ok

Elbot-Absolutely

Kush-Favorite Artist?

Elbot-Answering this question would violate my principles.

**Results**

A few minutes of conversation with each of these Artificial Intelligence Chatbots clearly displays the lack of depth in the conversation. They lack:

1.Interpretational skills.

2.Any form of memory- Each input statement is a new statement for them, irrespective of the input they have received before hand.

3.Cognitive skills

Kush Agrawal 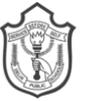

However,

Most of the bots were able to:

1.Answer simple/complex math problems.

2.Carry out their pre-defined objective.

**Discussion/Conclusion**

The above conversations clearly indicate that implementing a few set of skills into Machines such as calculation, answering with specific replies when given a specific input are not that hard to carry out. However when it comes to abstractions from the usual directive of the bots/machines, giving any form of sensible output becomes hard. Hence we conclude that as Moravec rightly said that it is the conscientious actions that are harder to implement in Artificial Intelligence, than the ones that require logic.

**Citations/References**

http://en.wikipedia.org/wiki/Rodney_Brooks

http://en.wikipedia.org/wiki/Marvin_Minsky

http://en.wikipedia.org/wiki/Hans_Moravec

http://en.wikipedia.org/wiki/Moravec's_paradox

http://highered.blogspot.com/2009/04/more-thoughts-on-moravecs-paradox.html

http://highered.blogspot.com/2009/03/learning-paradox.html

http://www.reddit.com/r/science/comments/88cfh/moravecs_paradox/

http://www2.cs.uregina.ca/~yyao/PAPERS/ieee_grc_08.pdf

http://www.frc.ri.cmu.edu/~hpm/project.archive/general.articles/1998/SimConEx.98.html

http://www.frc.ri.cmu.edu/~hpm/project.archive/robot.papers/2000/Cerebrum.html

http://www.frc.ri.cmu.edu/~hpm/project.archive/robot.papers/2000/puddle.html

http://www.frc.ri.cmu.edu/~hpm/project.archive/robot.papers/1999/SciAm.scan.html

http://www.frc.ri.cmu.edu/~hpm/project.archive/general.articles/1993/Robot93.html

http://www.transhumanist.com/volume1/moravec.htm

http://www.scientificamerican.com/article.cfm?id=rise-of-the-robots